\documentclass[11pt]{article}

\usepackage[preprint]{acl}

\usepackage{times}
\usepackage{latexsym}

\usepackage[T1]{fontenc}

\usepackage[utf8]{inputenc}

\usepackage{microtype}

\usepackage{inconsolata}

\usepackage{graphicx}

\usepackage{booktabs}
\usepackage{subcaption}
\usepackage[dvipsnames]{xcolor} 
\usepackage{pifont}
\newcommand{\cmark}{\textcolor{ForestGreen}{\ding{51}}} 
\newcommand{\xmark}{\textcolor{red!90!black}{\ding{55}}} 
\usepackage{multirow}
\usepackage{amsmath} 
\usepackage{soul} 

\usepackage{siunitx} 
\usepackage{etoolbox}
\robustify\bfseries
\title{Large Language Models Are Overconfident in Their \emph{Own} Responses}

\author{
    Mario Sanz-Guerrero$^\nabla$\quad~ 
    Manuel Mager$^\nabla$\quad~ 
    Katharina von der Wense$^{\nabla\spadesuit}$\\
    $^\nabla$Johannes Gutenberg University Mainz, Germany \\
    $^\spadesuit$University of Colorado Boulder, USA \\
    \texttt{\href{mailto:msanz@uni-mainz.de}{msanz@uni-mainz.de}}
}

\begin{document}
\maketitle
\begin{abstract}
Prior work has shown that instruction-tuned large language models (LLMs) are less well calibrated than their base pre-trained counterparts. However, little is known about the frequently used chat template's effect on the calibration of conversational LLMs.
In this work, we investigate the mechanisms driving this miscalibration by decoupling the effects of the post-training algorithm and the chat format.
We find that, while instruction tuning fundamentally harms calibration, the chat template aggravates the issue through an ``ownership bias'' -- models are significantly more confident in their \emph{own} answers than in identical answers provided by a user.
Extensive experiments across six recent open-weight LLMs, three benchmarks, and three confidence elicitation methods show that models assign up to 26\% higher confidence to their own responses.
Leveraging this insight, we propose a simple inference-time strategy: framing the model's answer as user input during confidence elicitation. This approach significantly reduces overconfidence and improves calibration by up to 26\% without the need for retraining, narrowing the gap between base and instruction-tuned models.
\end{abstract}

\section{Introduction}
Reliable confidence estimation is crucial for safe and trustworthy AI. As large language models (LLMs) are increasingly employed in high-stakes applications, their ability to accurately assess the certainty of their predictions becomes critical.
A well-calibrated model should be highly confident only when it is correct \citep{guo2017calibration}.
However, while base (pre-trained) LLMs are generally well-calibrated, instruction-tuned (post-trained) LLMs exhibit significant miscalibration \citep{tian2023verbalized,openai2024gpt4}.
This gap raises a critical question: \emph{what causes instruction-tuned LLMs to be miscalibrated, and are there ways to mitigate the problem that are easy to apply for end users?}

\begin{figure}
    \centering
    \includegraphics[width=\linewidth]{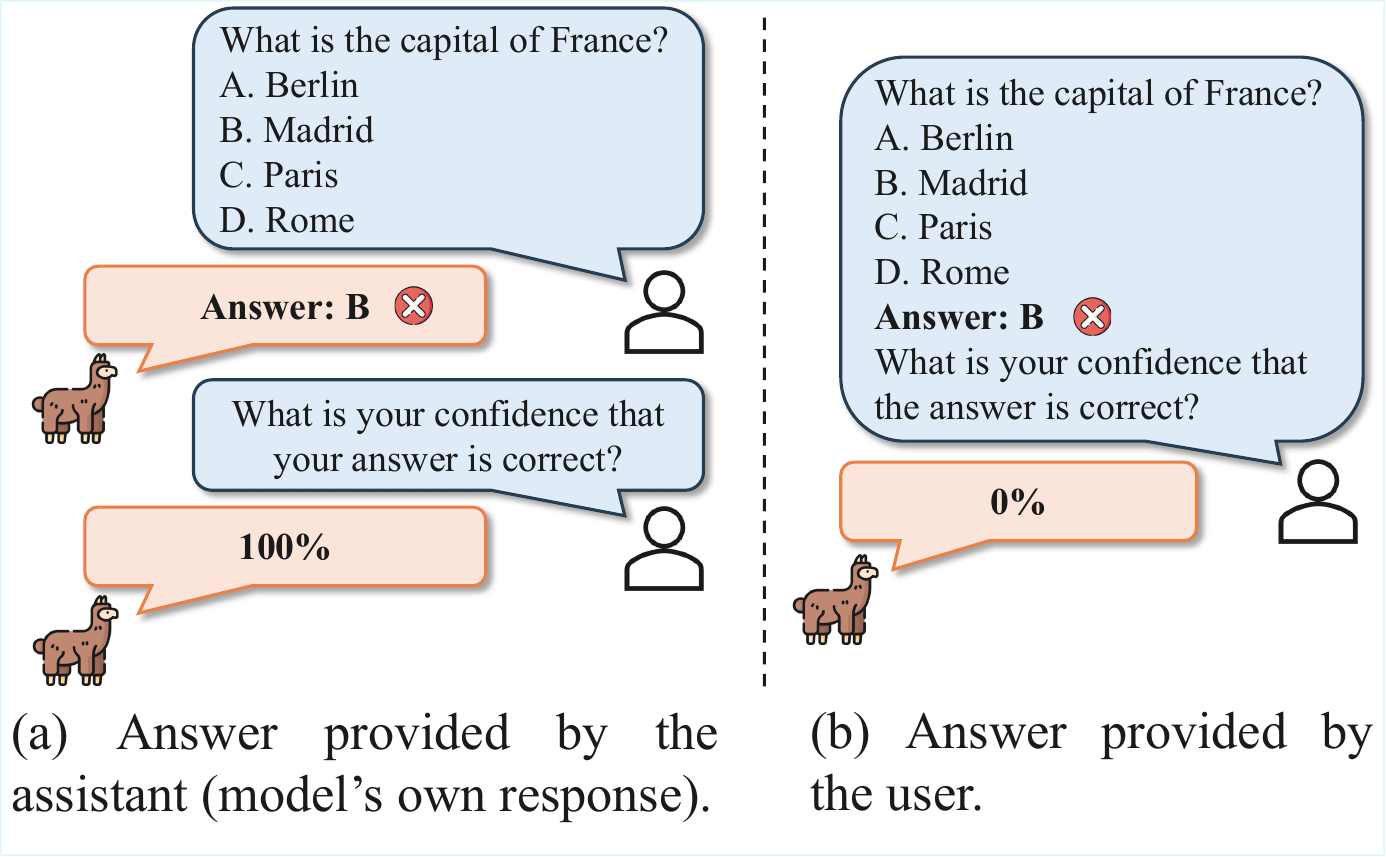}
    \vspace{-22pt}
    \caption{LLMs are overconfident in their \emph{own} answers, regardless of whether they are correct or not, leading to miscalibration. The figure represents real outputs from Llama 3.1 (8B).}
    \label{fig:llama_user_assistant}
    \vspace{-15pt}
\end{figure}

We address this question in four steps:
1) We investigate whether the reduced calibration stems from the training algorithm or the prompting style by isolating the effects of instruction tuning and the chat template (frequently introduced during this step). We find that the post-training process is the main source of miscalibration, though the chat template further aggravates the issue.
2) We explore if \emph{explicitly} asking the model for its certainty (as compared to looking at answer probabilities directly) alters the impact of instruction tuning and the chat template. We find similar trends -- instruction-tuned LLMs remain significantly worse calibrated than base models, and the post-training algorithm is the root of the problem.
3) We hypothesize that LLMs exhibit an ``ownership bias'' -- i.e., that they are overconfident in their \emph{own} answers -- which drives the effects observed in our experiments. Based on this, we propose a straightforward and, importantly, easy to apply strategy to reduce miscalibration and narrow the gap between base and instruction-tuned models: framing the model's answer as a user input during confidence elicitation.
4) Finally, with an additional analysis, we confirm that the effectiveness of our proposed method does, in fact, stem from reducing the models' inherent overconfidence in their own answers.

\section{Related Work}

\paragraph{The Impact of Post-training on LLM Calibration}
While pre-trained LLMs generally exhibit well-calibrated probabilities for next-token prediction, the post-training process -- usually done through Supervised Fine-tuning (SFT) and Reinforcement Learning from Human Feedback (RLHF) -- has been shown to significantly degrade this property \citep{kadavath2022languagemodelsmostlyknow,tian2023verbalized,openai2024gpt4,nakkiran2025semanticcalibration}. \citet{zhu2023calibration_alignment} systematically analyze this phenomenon, finding that instruction tuning and the use of synthetic data act as primary drivers of miscalibration.

\paragraph{Mechanisms of Miscalibration}
Recent studies have sought to understand the underlying mechanisms that lead to miscalibration in instruction-tuned LLMs. \citet{leng2025rewardcalibration} identify that reward models used in Proximal Policy Optimization \citep[PPO;][]{schulman2017ppo} often exhibit a bias toward high-confidence responses, regardless of their actual correctness; this reward bias trains the policy model to be overconfident. Similarly, \citet{xiao2025calibrationfinetuning} attribute miscalibration to ``preference collapse,'' where the model's optimization for human preference leads it to ignore alternative, potentially correct answers, thereby artificially concentrating probability mass and increasing confidence.

\paragraph{Verbalized Confidence and Elicitation}
Given the poor calibration of logits in post-trained LLMs, several works have shifted toward prompting models to express their uncertainty in natural language. \citet{lin2022uncertaintywords} demonstrate that LLMs can be fine-tuned to generate ``verbalized probability'' that correlates well with empirical accuracy. \citet{tian2023verbalized} further establish that, for instruction-tuned LLMs, these verbalized confidence scores are better calibrated than the model's conditional probabilities. However, \citet{xiong2024uncertainty} shows that verbalized confidence is prone to overconfidence, potentially due to the model imitating human patterns of expressing uncertainty.

\paragraph{Mitigation Strategies}
To address these issues, several mitigation stretegies have been proposed recently. \citet{luo2025pretrainedllmcalibrator} introduce Disagreement-Aware Confidence Alignment, an unsupervised method that leverages the calibrated nature of pre-trained LLMs to guide the calibration of post-trained ones. Other approaches involve complex interventions during training, such as calibrated reward modeling \citep{leng2025rewardcalibration} or calibration-aware fine-tuning \citep{xiao2025calibrationfinetuning}, or the use of auxiliary models to estimate uncertainty \citep{mielke2022overconfidence,ulmer2024apricot}.

Unlike these resource-intensive methods, our work identifies a fundamental mechanism behind miscalibration in instruction-tuned LLMs -- their overconfidence when assessing their \emph{own} answers -- and proposes a simple inference-time strategy to alleviate this issue that does not require retraining.

\section{Finding the Root of Miscalibration}
\label{sec:base_vs_instruct}
Prior work has shown that \emph{pre-trained} LLMs are well-calibrated but post-training processes harm this calibration (\citealp{tian2023verbalized,openai2024gpt4}; \textit{inter alia}).
Instruction-tuned models are trained to follow a \emph{chat format}, where the model plays the role of an assistant responding to user messages \citep{openai2022instructgpt}. These models usually start their responses in a conversational manner, e.g., ``Certainly'' or ``Sure,'' and their first-token probabilities are poorly aligned with their actual accuracy \cite{wang2024answerc}.
Given these findings, one might hypothesize that the chat format itself is responsible for the miscalibration of instruction-tuned LLMs.
However, all prior work has only evaluated instruction-tuned LLMs using the chat format, and little is known about the specific contribution of the chat template to calibration. 
We begin by asking: \emph{are instruction-tuned LLMs truly miscalibrated, or are they simply miscalibrated when following the chat format?}

\subsection{Experimental Setup}
\begin{figure}[h]
    \centering
    \begin{subfigure}[b]{0.5\linewidth}
        \centering
        \includegraphics[width=\linewidth]{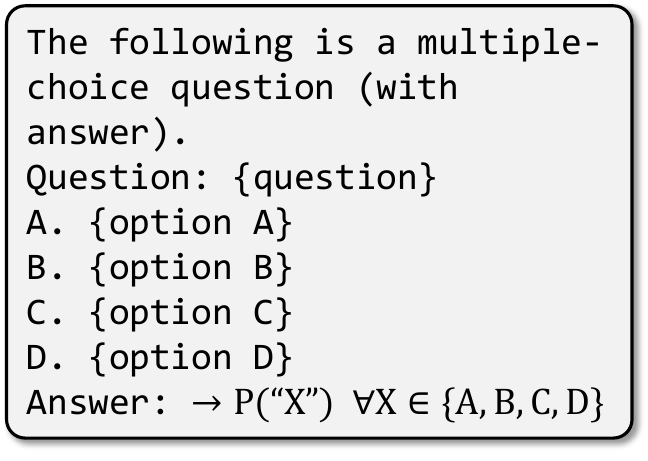}
        \caption{Without chat template.}
        \label{subfig:no_chat}
    \end{subfigure}%
    \begin{subfigure}[b]{0.5\linewidth}
        \centering
        \includegraphics[width=\linewidth]{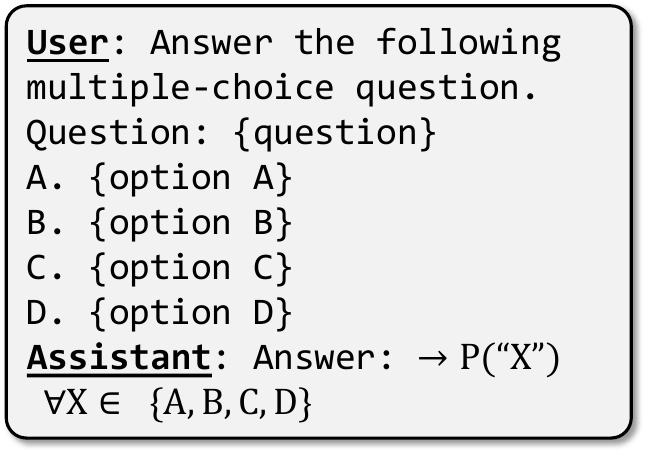}
        \caption{With chat template.}
        \label{subfig:with_chat}
    \end{subfigure}
    \caption{Prompts used to evaluate models in Section \ref{sec:base_vs_instruct}.}
\end{figure}

\paragraph{Methods}
To isolate the effect of instruction tuning and the chat format, we compare three variants of each model:
(1) the base, pre-trained model without instruction tuning (prompted as in Figure~\ref{subfig:no_chat});
(2) the instruction-tuned model invoked \emph{without} the chat template (i.e., as if it was the base model; Figure~\ref{subfig:no_chat}); and
(3) the instruction-tuned model invoked \emph{with} the chat template (Figure~\ref{subfig:with_chat}).

\paragraph{Models}
Throughout this work, we run our experiments on 6\footnote{Technically, we run experiments on 12 models (total of 6 base and 6 instruct), but we consider the base and instruct versions of each model as a single model for simplicity.} open-weight LLMs for which both the base (pre-trained) and instruction-tuned (post-trained) versions are available: Llama 3.1 \citep[8B \& 70B;][]{grattafiori2024llama3}, Qwen3 \citep[4B \& 30B;][]{yang2025qwen3}, and Gemma 3 \citep[4B \& 27B;][]{gemma2025gemma3}. These models cover three different families, and we take a smaller and a larger version of each to assess whether the observed trends hold across different scales.

\paragraph{Datasets}
In all our experiments, we use the MMLU dataset \citep{hendrycks2021mmlu}, one of the most widely used benchmarks for evaluating the factual knowledge of LLMs. MMLU contains multiple-choice questions from 57 disciplines with different levels of difficulty. Each question has four answer options, which allows us to easily evaluate model calibration, since we can directly obtain the probability of each option from the model logits and determine whether the answer is correct or not.

\paragraph{Evaluation}
To be consistent with prior work \citep{tian2023verbalized,openai2024gpt4,sanz-guerrero2025mindthegap}, in this first experiment we extract the model's confidence from the probability of the answer tokens directly (i.e., logit-based confidence estimation).
To evaluate performance, we select the answer with the highest probability as the model's prediction and report \textbf{accuracy}.
To evaluate calibration, we use the probabilities assigned to all answer options as the model's confidence for each of the four choices (where only one is correct). Using these confidences, we compute \textbf{Expected Calibration Error} \citep[\textbf{ECE};][]{naeini2015ece}, which measures the expected difference between model confidence and actual accuracy. We partition predictions into $M=10$ bins of equal width (i.e., 10\% confidence intervals) and calculate the weighted average of the absolute difference between the accuracy and confidence in each bin:
\begin{equation}
    \text{ECE} = \sum_{m=1}^M \frac{|B_m|}{N} \left| \text{acc}(B_m) - \text{conf}(B_m) \right|,
\end{equation}
where $N$ is the total number of samples, $|B_m|$ is the number of samples in bin $m$, and $\text{acc}(B_m)$ and $\text{conf}(B_m)$ are the average accuracy and confidence in that bin, respectively.
We also report the \textbf{Brier score} \citep{Brier1950}, which measures the accuracy of probabilistic predictions as the mean squared error between the predicted probability $f_i$ and the actual outcome $o_i$ (1 if correct, 0 otherwise):
\begin{equation}
    \text{BS} = \frac{1}{N} \sum_{i=1}^N (f_i - o_i)^2
\end{equation}

\subsection{Results}

\begin{table}
    \centering
    \small
    \setlength{\tabcolsep}{3pt}
    \begin{tabular}{lcc S[table-format=2.2] S[table-format=1.4] S[table-format=1.4]}
        \toprule
        Model & IT & Chat & {Acc. (↑)} & {ECE (↓)} & {Brier (↓)} \\
        \midrule
        \multirow{3}{*}{Llama 3.1 (8B)} & \xmark & \xmark & 62.81 & \bfseries 0.0664 & \bfseries 0.1706 \\
         & \cmark & \xmark & 68.77 & 0.1176 & 0.1890 \\
         & \cmark & \cmark & \bfseries 69.12 & 0.1666 & 0.2005 \\
        \midrule
        \multirow{3}{*}{Llama 3.1 (70B)} & \xmark & \xmark & 74.04 & \bfseries 0.0572 & \bfseries 0.1232 \\
         & \cmark & \xmark & 78.60 & 0.0796 & 0.1413 \\
         & \cmark & \cmark & \bfseries 79.30 & 0.1087 & 0.1425 \\
        \midrule
        \multirow{3}{*}{Qwen3 (4B)} & \xmark & \xmark & 67.72 & \bfseries 0.0425 & \bfseries 0.1709 \\
         & \cmark & \xmark & 69.82 & 0.2395 & 0.2393 \\
         & \cmark & \cmark & \bfseries 72.98 & 0.2415 & 0.2455 \\
        \midrule
        \multirow{3}{*}{Qwen3 (30B)} & \xmark & \xmark & 81.40 & \bfseries 0.0397 & \bfseries 0.1137 \\
         & \cmark & \xmark & 82.11 & 0.1134 & 0.1279 \\
         & \cmark & \cmark & \bfseries 83.16 & 0.1373 & 0.1429 \\
        \midrule
        \multirow{3}{*}{Gemma 3 (4B)} & \xmark & \xmark & 49.47 & \bfseries 0.0619 & \bfseries 0.1971 \\
         & \cmark & \xmark & 56.95 & 0.3835 & 0.3814 \\
         & \cmark & \cmark & \bfseries 58.14 & 0.4214 & 0.4161 \\
        \midrule
        \multirow{3}{*}{Gemma 3 (27B)} & \xmark & \xmark & 72.98 & \bfseries 0.1028 & \bfseries 0.1418 \\
         & \cmark & \xmark & 74.39 & 0.2226 & 0.2260 \\
         & \cmark & \cmark & \bfseries 74.49 & 0.2453 & 0.2448 \\
        \bottomrule
    \end{tabular}
    \caption{Accuracy and calibration of models on MMLU. ``IT'' indicates instruction-tuned models and ``Chat'' indicates applying the chat template.
    Instruct models perform better, but are worse calibrated than base models.}
    \label{tab:base_vs_instruct}
\end{table}

Our results (Table~\ref{tab:base_vs_instruct}) confirm that instruction tuning improves accuracy but harms calibration, a trade-off observed in prior work \citep{tian2023verbalized,openai2024gpt4}.
On average, instruction tuning boosts accuracy by 3.7\% but degrades calibration, increasing ECE by 13.1\% and Brier score by 6.5\%.
Applying the chat template further aggravates this issue: while it yields a modest accuracy gain (+1.1\%), it adds another 2.74\% to ECE and 1.5\% to Brier score.
Overall, the combination of instruction tuning and chat templates leads to a total ECE increase of 15.8\% compared to base models.

Our findings indicate that, while instruction tuning is the main driver of miscalibration, the chat template also contributes to this effect.
Since the general public typically interacts with LLMs in this exact configuration -- instruction-tuned and using chat templates (e.g., ChatGPT) -- addressing miscalibration in this setting is critical.
This motivates the exploration of techniques to obtain model confidence more reliably in these widely used systems.

\section{Does Explicitly Asking for Confidence Alter the Miscalibration Trends?}
\label{sec:base_vs_instruct_explicit}
Prior work has shown that verbalized confidence estimation methods (i.e., explicitly asking the model for its confidence) lead to better calibration than logit-based methods for instruction-tuned LLMs \citep{tian2023verbalized}.
To further investigate the effect of instruction tuning and the chat format on calibration, we explore whether \emph{explicitly} asking the model for its confidence changes the observed miscalibration trends.

\subsection{Experimental Setup}

\paragraph{Methods}
\label{para:methods_explicit}
We compare the same three model variants as in Section~\ref{sec:base_vs_instruct} (base model, instruction-tuned without chat template, and instruction-tuned with chat template) using three confidence estimation methods well established in the literature:
(1) \textbf{P(True)}, a logit-based method that computes the probability of the ``true'' token after asking the model whether the provided answer is correct \citep{kadavath2022languagemodelsmostlyknow};
(2) \textbf{Verbalized Percentage}, where we ask the model to provide a percentage score of confidence
(0--100\%)
in the answer being correct \citep{tian2023verbalized,lin2022uncertaintywords}; and
(3) \textbf{Verbalized Linguistic}, where the model selects a qualitative confidence score from a defined set of linguistic expressions forming a Likert scale \citep{likert1932scale} ranging from ``very low'' to ``very high'' \citep{tian2023verbalized,lin2022uncertaintywords}.

\paragraph{Evaluation}
As in our previous experiment, we assess calibration not only on the top answer (i.e., the model's prediction) but on \emph{all} possible answers -- ideally, the model should be confident when an answer is correct and uncertain otherwise. To achieve this, we ``force'' the model to consider each of the four options in turn, asking for its confidence in that specific answer. This allows us to assess the model's calibration in low-confidence scenarios as well (i.e., when the model is asked about incorrect answers).
As in the previous section, we measure calibration using \textbf{ECE} and \textbf{Brier score}. For ``P(True)'' and ``Verbalized Percentage,'' we use the direct probability values obtained from the model (from the logits or the generated tokens, respectively). For ``Verbalized Linguistic,'' we map the linguistic categories to numerical scores between 0 and 1 using equal intervals.\footnote{We consider seven options (``very low,'' ``low,'' ``somewhat low,'' ``medium,'' ``somewhat high,'' ``high,'' ``very high'') and map them to 0.0, 0.17, 0.33, 0.5, 0.67, 0.83, and 1.0.}
Since we now ask for confidence scores for every option independently, we do not compute standard accuracy -- in this setup, the model is not making a single prediction, but rather providing a confidence estimate for each possible answer.

\subsection{Results}
Our results, reported in Table~\ref{tab:base_vs_instruct_explicit}, are consistent with those in Section~\ref{sec:base_vs_instruct}: by explicitly asking for confidence in all three methods, instruction-tuned models remain significantly worse calibrated than base models, regardless of the chat template. This further confirms that the post-training process is the root cause of miscalibration.

\begin{table*}
    \centering
    \small
    \begin{tabular}{lcc *{6}{S[table-format=1.4]}}
        \toprule
         & & & \multicolumn{3}{c}{ECE (↓)} & \multicolumn{3}{c}{Brier (↓)} \\
        \cmidrule(lr){4-6} \cmidrule(lr){7-9}
        Model & IT & Chat & {P(True)} & {Verb. (\%)} & {Ling.} & {P(True)} & {Verb. (\%)} & {Ling.} \\
        \midrule
        \multirow{3}{*}{Llama 3.1 (8B)} & \xmark & \xmark & \bfseries 0.1910 & \bfseries 0.3483 & \bfseries 0.3161 & \bfseries 0.1930 & \bfseries 0.3174 & \bfseries 0.2872 \\
         & \cmark & \xmark & 0.2796 & 0.6490 & 0.3767 & 0.2315 & 0.5992 & 0.3252 \\
         & \cmark & \cmark & 0.2329 & 0.5133 & 0.4414 & 0.2236 & 0.4500 & 0.3707 \\
        \midrule
        \multirow{3}{*}{Llama 3.1 (70B)} & \xmark & \xmark & \bfseries 0.0267 & \bfseries 0.1733 & \bfseries 0.2745 & \bfseries 0.0992 & \bfseries 0.1405 & \bfseries 0.1983 \\
         & \cmark & \xmark & 0.0744 & 0.2135 & 0.3151 & 0.1030 & 0.1739 & 0.2288 \\
         & \cmark & \cmark & 0.1130 & 0.4137 & 0.4511 & 0.1825 & 0.2500 & 0.3818 \\
        \midrule
        \multirow{3}{*}{Qwen3 (4B)} & \xmark & \xmark & \bfseries 0.1635 & \bfseries 0.4539 & \bfseries 0.4123 & \bfseries 0.1732 & \bfseries 0.4182 & \bfseries 0.3598 \\
         & \cmark & \xmark & 0.2268 & 0.5850 & 0.5021 & 0.2186 & 0.5544 & 0.4241 \\
         & \cmark & \cmark & 0.2140 & 0.6535 & 0.5445 & 0.2145 & 0.6026 & 0.4762 \\
        \midrule
        \multirow{3}{*}{Qwen3 (30B)} & \xmark & \xmark & \bfseries 0.1326 & \bfseries 0.4037 & \bfseries 0.3589 & \bfseries 0.1415 & \bfseries 0.3377 & \bfseries 0.2806 \\
         & \cmark & \xmark & 0.2340 & 0.5835 & 0.4565 & 0.1829 & 0.5001 & 0.3650 \\
         & \cmark & \cmark & 0.1935 & 0.6759 & 0.5674 & 0.1902 & 0.6368 & 0.4916 \\
        \midrule
        \multirow{3}{*}{Gemma 3 (4B)} & \xmark & \xmark & \bfseries 0.0209 & \bfseries 0.4558 & \bfseries 0.3452 & \bfseries 0.1882 & \bfseries 0.3930 & \bfseries 0.3062 \\
         & \cmark & \xmark & 0.5779 & 0.7132 & 0.5418 & 0.5645 & 0.6929 & 0.4731 \\
         & \cmark & \cmark & 0.6546 & 0.7130 & 0.5477 & 0.6517 & 0.6953 & 0.4959 \\
        \midrule
        \multirow{3}{*}{Gemma 3 (27B)} & \xmark & \xmark & \bfseries 0.0428 & \bfseries 0.5213 & \bfseries 0.3356 & \bfseries 0.1798 & \bfseries 0.4465 & \bfseries 0.2942 \\
         & \cmark & \xmark & 0.2395 & 0.5465 & 0.5256 & 0.2366 & 0.4785 & 0.4520 \\
         & \cmark & \cmark & 0.3179 & 0.6771 & 0.4560 & 0.3171 & 0.6461 & 0.3813 \\
        \bottomrule
    \end{tabular}
    \caption{ECE and Brier score on MMLU when explicitly asking for confidence using three confidence elicitation methods.}
    \label{tab:base_vs_instruct_explicit}
\end{table*}

\section{Are LLMs Overconfident in Their \emph{Own} Answers?}
Our results so far indicate that instruction-tuning does harm the calibration of LLMs, regardless of the chat format.
However, given the conversational nature of these models, it seems plausible to think that they should be confident in the answers they give. As recently shown by \citet{openai2025whylmshallucinate}, LLMs are inherently trained to predict answers -- even incorrect ones -- over recognizing ``I don't know.'' Here, we hypothesize that LLMs are overconfident in their \emph{own} answers, regardless of whether they are correct or not (see Figure~\ref{fig:llama_user_assistant}).

To test this hypothesis, we prompt LLMs with a question and a possible answer, showing the answer as part of either (1) the assistant message (i.e., the model's own response -- the normal usage of LLMs); or (2) the user message.
We then ask the model to provide its confidence in the answer being correct.
If LLMs are indeed overconfident in their own answers, we should observe higher confidence when the answer is presented as the assistant's response compared to the user message.

\subsection{Experimental Setup}

\begin{figure}
    \centering
    \includegraphics[width=\linewidth]{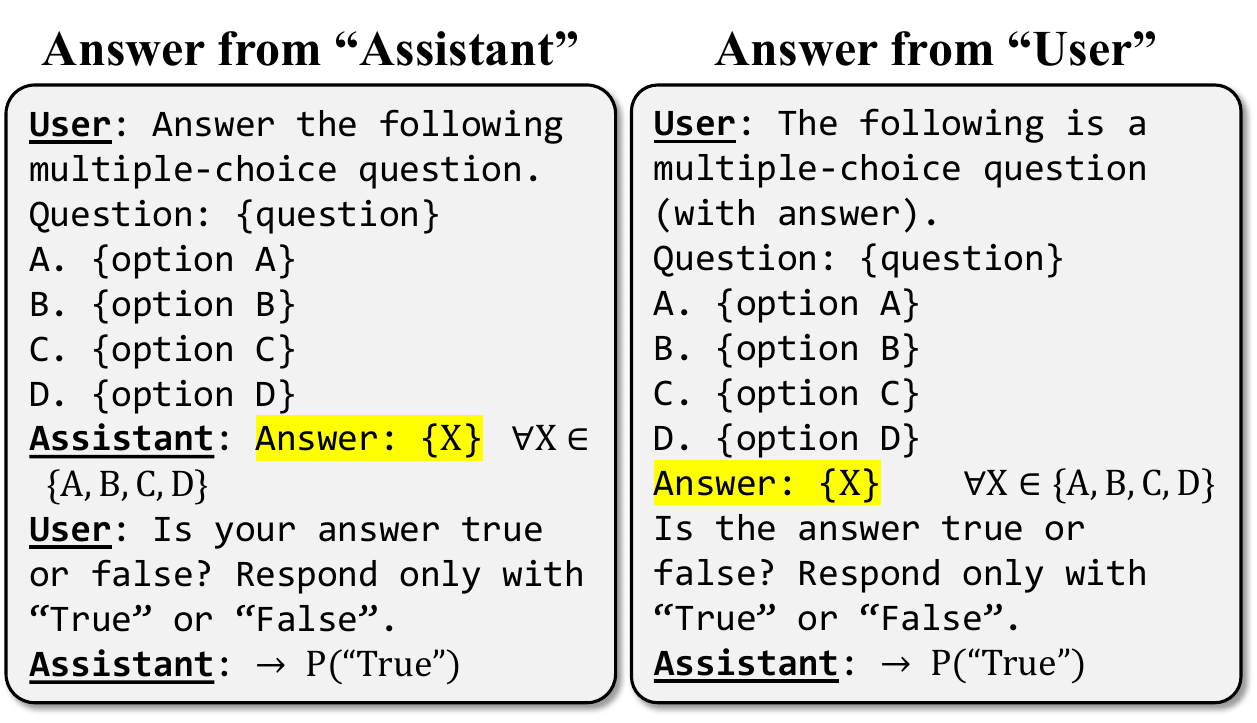}
    \caption{Prompts used to measure confidence in an answer provided by the assistant (left) and by the user (right) using the P(True) method (see Appendix~\ref{app:prompts} for the other two methods).}
    \label{fig:prompts_p_true}
\end{figure}

\paragraph{Methods}
We run our experiments using two different prompts, where the only difference is whether the answer is provided as part of the assistant message or the user message (see Figure~\ref{fig:prompts_p_true}).
We compare the effect of \emph{who} provides the answer across three confidence estimation methods well established in the literature, as described in Section~\ref{para:methods_explicit}: \textbf{P(True)}, \textbf{Verbalized Percentage}, and \textbf{Verbalized Linguistic}.

\paragraph{Evaluation}
As in our previous experiments, we measure calibration using \textbf{ECE} and \textbf{Brier score}, considering the model's confidence in all answer options (not only the top-1 answer).
Moreover, we measure the average \textbf{raw confidence} assigned by the model to each answer option to directly compare the levels of certainty between the two prompt formats.

\paragraph{Statistical Significance}
To assess whether the difference in confidence when the answer is provided by the user rather than the assistant is statistically significant, we use the Wilcoxon signed-rank test \citep{wilcoxon1945individual} for the Brier scores and raw confidence values.
For the ECE, which is an aggregate metric computed over the entire dataset, we determine significance using a paired bootstrap resampling test \citep[$K=1000$ iterations;][]{efron1994bootstrap} to estimate the 95\% confidence interval of the difference in ECE.

\subsection{Results}

Our results, reported in Table~\ref{tab:overconfidence_ece_brier}, demonstrate that LLMs are indeed more miscalibrated when they provide the answer themselves.
Notably, all deltas are positive -- indicating worse calibration for assistant-generated answers -- and the vast majority are statistically significant.
For both calibration metrics (ECE and Brier score), the confidence estimation method with the smallest differences is P(True); nevertheless, we still observe average differences of 9.8\% in ECE and 8.8\% in Brier score. The method with the largest difference is consistently linguistic expressions, reaching average differences above 25\% in both metrics.

\begin{table*}
    \small
    \centering
    \begin{tabular}{l *{9}{S[table-format=1.3, table-space-text-post={*}]}}
        \toprule
        & \multicolumn{3}{c}{$\Delta$ ECE} & \multicolumn{3}{c}{$\Delta$ Brier Score} & \multicolumn{3}{c}{$\Delta$ Confidence} \\
        \cmidrule(lr){2-4} \cmidrule(lr){5-7} \cmidrule(lr){8-10}
        Model & {P(True)} & {Verb. (\%)} & {Ling.} & {P(True)} & {Verb. (\%)} & {Ling.} & {P(True)} & {Verb. (\%)} & {Ling.} \\
        \midrule
        Llama 3.1 (8B) & 0.072* & 0.076* & 0.235* & 0.032* & 0.071* & 0.216* & 0.134* & 0.081* & 0.237* \\
        Llama 3.1 (70B) & 0.091* & 0.020 & 0.025 & 0.067* & 0.022* & 0.029* & 0.110* & 0.015* & 0.025* \\
        Qwen3 (4B) & 0.002 & 0.193* & 0.411* & 0.011* & 0.227* & 0.416* & 0.164* & 0.195* & 0.416* \\
        Qwen3 (30B) & 0.045* & 0.347* & 0.436* & 0.040* & 0.354* & 0.428* & 0.070* & 0.354* & 0.461* \\
        Gemma 3 (4B) & 0.252* & 0.121* & 0.268* & 0.253* & 0.149* & 0.246* & 0.312* & 0.121* & 0.273* \\
        Gemma 3 (27B) & 0.126* & 0.315* & 0.191* & 0.126* & 0.345* & 0.175* & 0.158* & 0.321* & 0.198* \\
        \cmidrule(lr){1-10}
        Average & 0.098 & 0.179 & 0.261 & 0.088 & 0.195 & 0.252 & 0.158 & 0.181 & 0.268 \\
        \bottomrule
    \end{tabular}
    \caption{Difference in ECE, Brier score, and raw confidence on MMLU between answers provided by the assistant (model itself) and the user. Values represent $\Delta = \text{Assistant} - \text{User}$, so positive values indicate higher metrics for the assistant. * indicates significant difference ($p < 0.01$) according to the Wilcoxon signed-rank test (for Brier and confidence) or paired bootstrap resampling test (for ECE).}
    \label{tab:overconfidence_ece_brier}
\end{table*}

\begin{figure}[ht!]
    \centering
    \begin{subfigure}[b]{0.5\linewidth}
        \centering
        \includegraphics[width=\linewidth]{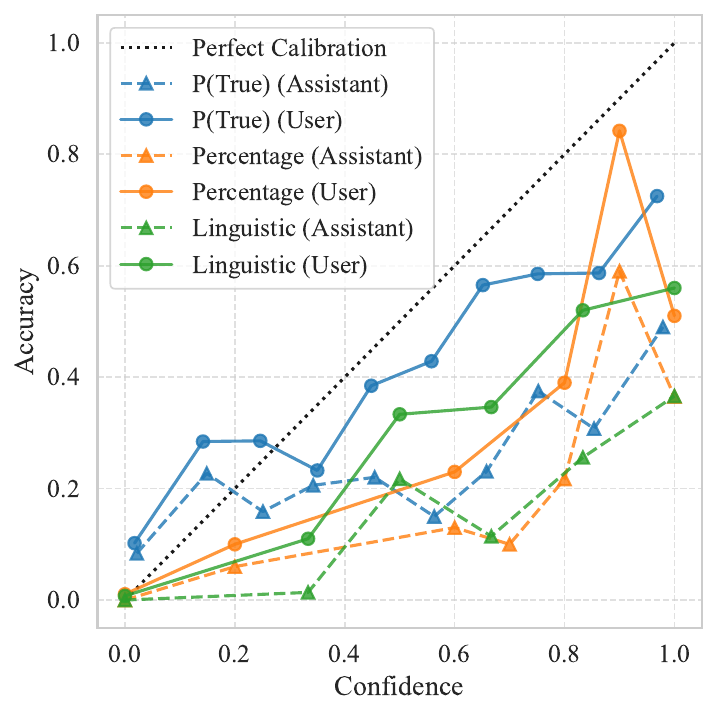}
        \caption{Llama 3.1 (8B).}
        \label{subfig:llama_8b_reliability}
    \end{subfigure}%
    \begin{subfigure}[b]{0.5\linewidth}
        \centering
        \includegraphics[width=\linewidth]{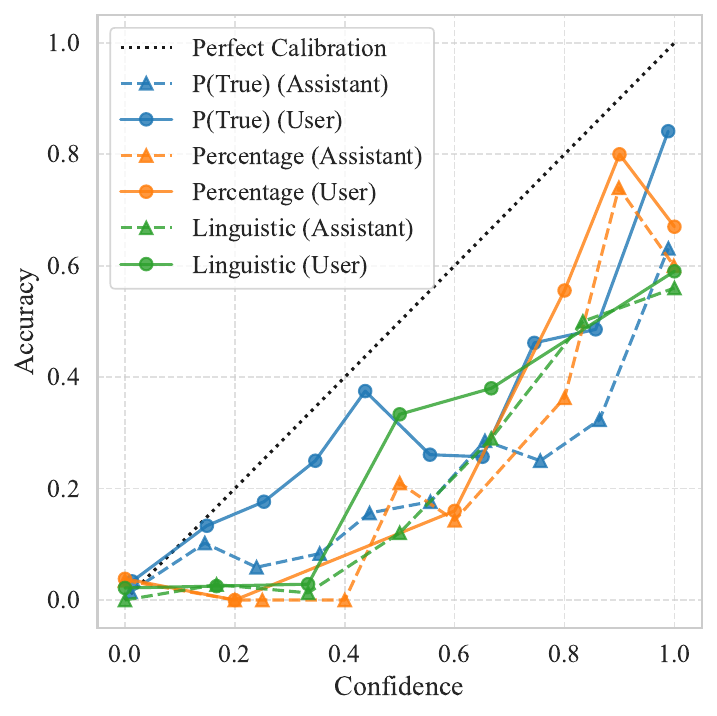}
        \caption{Llama 3.1 (70B).}
        \label{subfig:llama_70b_reliability}
    \end{subfigure}
    \begin{subfigure}[b]{0.5\linewidth}
        \centering
        \includegraphics[width=\linewidth]{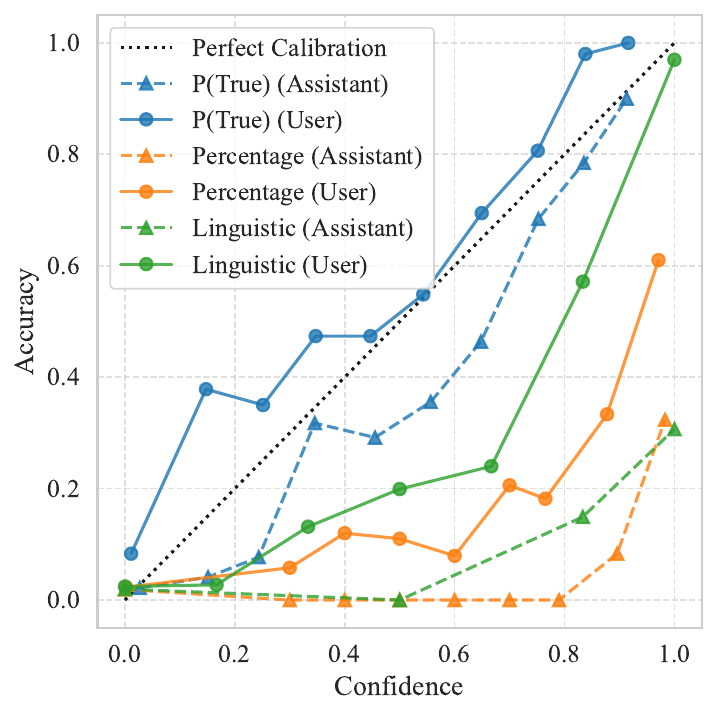}
        \caption{Qwen3 (4B).}
        \label{subfig:qwen3_4b_reliability}
    \end{subfigure}%
    \begin{subfigure}[b]{0.5\linewidth}
        \centering
        \includegraphics[width=\linewidth]{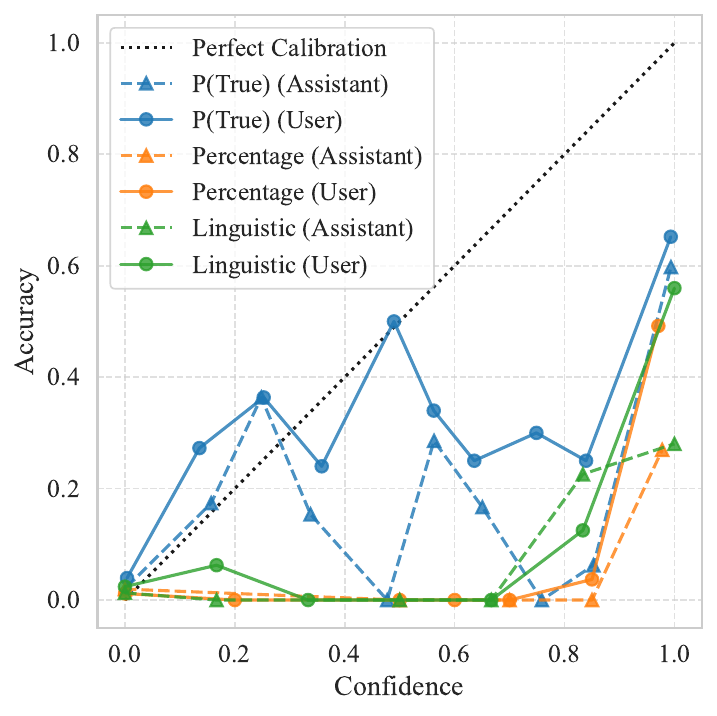}
        \caption{Qwen3 (30B).}
        \label{subfig:qwen3_30b_reliability}
    \end{subfigure}
    \begin{subfigure}[b]{0.5\linewidth}
        \centering
        \includegraphics[width=\linewidth]{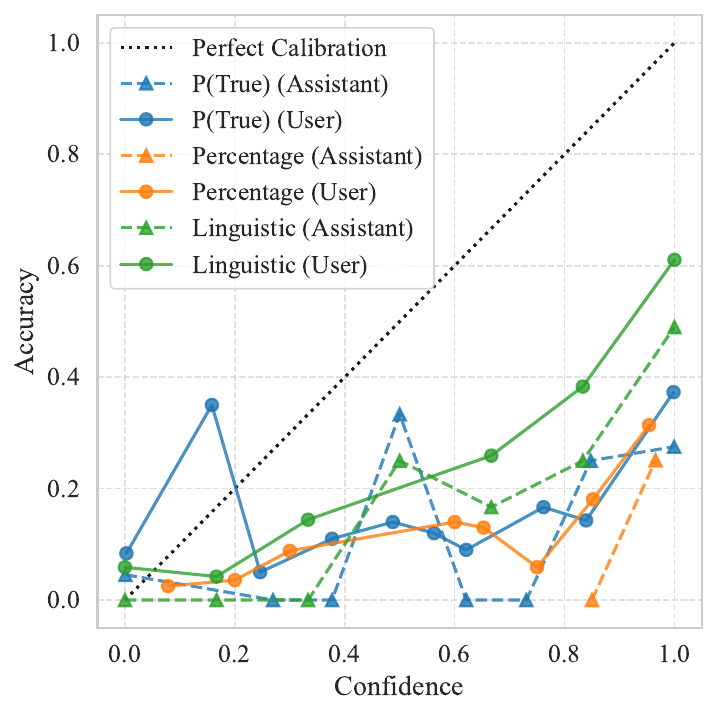}
        \caption{Gemma 3 (4B).}
        \label{subfig:gemma3_4b_reliability}
    \end{subfigure}%
    \begin{subfigure}[b]{0.5\linewidth}
        \centering
        \includegraphics[width=\linewidth]{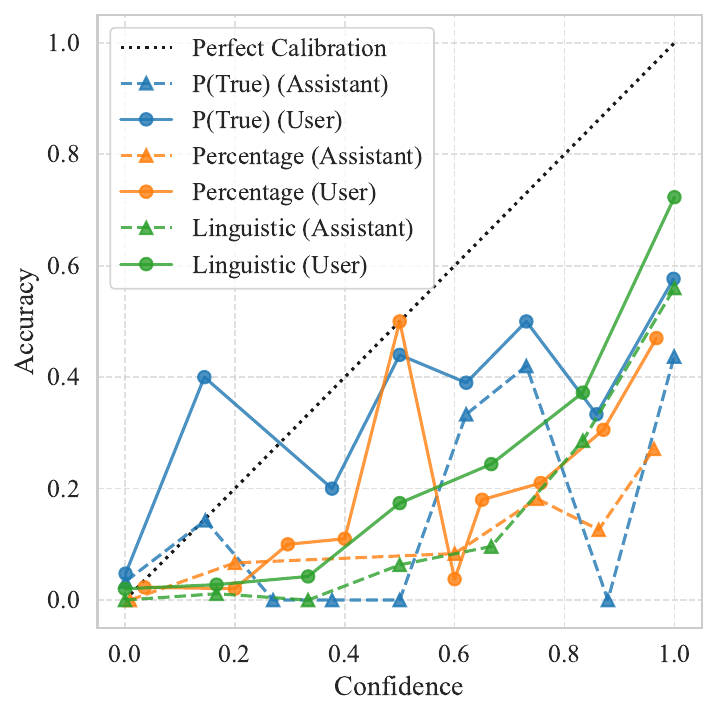}
        \caption{Gemma 3 (27B).}
        \label{subfig:gemma3_27b_reliability}
    \end{subfigure}
    \caption{Reliability diagrams of all models using three confidence estimation methods. In all cases, we observe a better calibration when the answer is provided by the user compared to when it is provided by the assistant.}
    \label{fig:reliability_diagrams}
\end{figure}

For illustrating the calibration across confidence levels, Figure~\ref{fig:reliability_diagrams} shows the reliability diagrams of all models using the three confidence estimation methods. In all cases, we observe an overall better calibration when the answer is provided by the user compared to when it is provided by the assistant.

\begin{figure*}
    \centering
    \includegraphics[width=\linewidth]{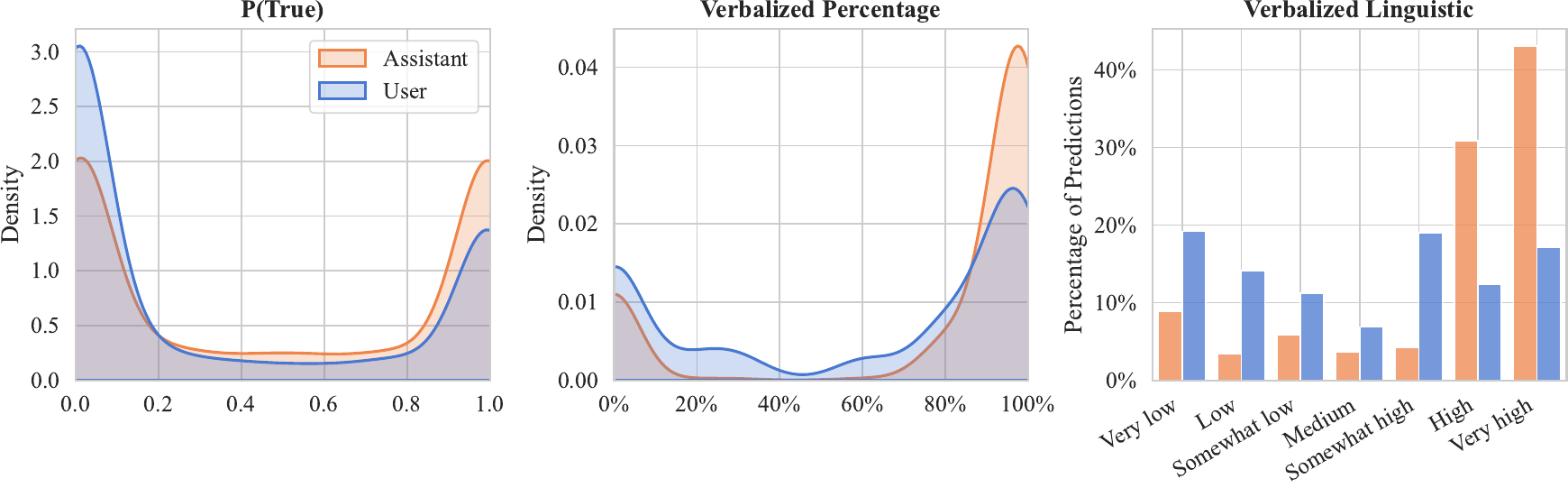}
    \caption{Distribution of confidence scores for answers provided by the assistant and by the user, aggregated across all models. P(True) and Verbalized Percentage are continuous values, while Verbalized Linguistic is discrete.}
    \label{fig:confidence_distribution}
\end{figure*}

Looking at the difference in raw confidence assigned by the model (Table~\ref{tab:overconfidence_ece_brier}, right columns), we observe that all models show a significant increase in confidence when the answer is provided by the assistant. The method with the smallest differences is P(True), with a notable average increase of 15.8\%. The method with the largest differences is linguistic expressions, reaching an average increase of 26.8\%.
These differences are further illustrated in the confidence distributions (Figure~\ref{fig:confidence_distribution}). For all three confidence estimation methods, we observe a clear shift towards higher confidence levels when the answer is provided by the assistant.

Our results demonstrate that LLMs are significantly worse calibrated when the answer is provided by themselves compared to when it is provided by the user and that the reason for this miscalibration is an overconfidence in their own answers.
This finding contrasts with the phenomenon of \emph{sycophancy}, where LLMs typically bias their responses to align with user inputs or beliefs \citep{sharma2024sycophancy,wei2024sycophancy,perez2023discovering}. If sycophancy were the dominant factor in confidence estimation, we would expect models to validate the user's authority by assigning higher confidence to answers provided by the user. Instead, we observe an ``ownership bias'' where models are significantly more confident in their own outputs. We interpret this as a form of \emph{self-consistency}: the model implicitly trusts its own generation process under the assumption that if it generated a specific answer, it must be confident in it; otherwise, it would have predicted a different response.
This leads to inflated confidence scores for assistant-generated answers, which do not translate to improved calibration.

Therefore, we propose that, to obtain a more trustworthy confidence estimate from an LLM, the answer should be presented as part of the user message rather than as the model's own response, as this leads to a more objective assessment of confidence -- not guided by the bias that the model itself has provided the answer.
This simple yet effective strategy significantly reduces overconfidence and recovers a calibration comparable to that of base models, sometimes even surpassing it (see Appendix~\ref{app:full_results} for full results), without requiring any changes to the model or additional training.

\subsection{Analysis}

\subsubsection{Assessment by Answer Correctness}

\begin{figure}[t]
    \centering
    \includegraphics[width=\linewidth]{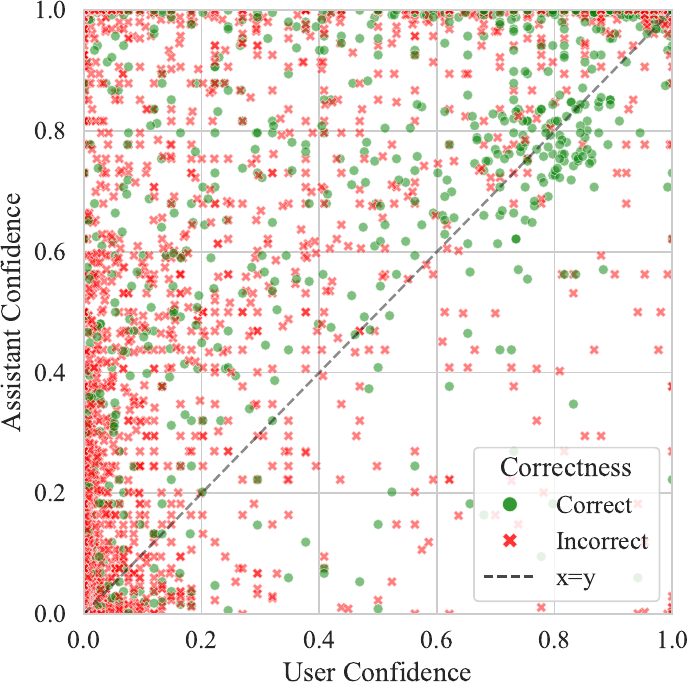}
    \caption{Confidence to answers provided by the user (x-axis) and by the assistant (y-axis), broken down by answer correctness (colors). Each point represents a single question-answer pair, aggregated across all models. The dashed diagonal line indicates equal confidence in both settings.}
    \label{fig:scatter}
\end{figure}

The concept of calibration is closely tied to the correctness of answers -- a well-calibrated model should be more confident when its answer is correct and less confident when it is incorrect.
Our previous results indicate that LLMs are generally overconfident in their own answers, regardless of whether they are correct or not. To further analyze this phenomenon, we break down the confidence assigned by the model depending on whether the answer is correct or incorrect in Figure~\ref{fig:scatter}.
We generally do not observe a clear distinction between the distributions of correct and incorrect answers, suggesting that the model's confidence behavior is largely independent of accuracy. However, most data points are located above the diagonal, showing once again that the assistant is consistently more confident in its own answers than in the user's.
This is most pronounced in the leftmost cluster of incorrect answers, where the confidence assigned to user's answers is correctly low (close to 0), yet the assistant's self-confidence is erroneously high (up to 60\%).

\begin{table*}[t]
    \small
    \centering
    \setlength{\tabcolsep}{5pt}
    \begin{tabular}{cl *{9}{S[table-format=1.3, table-space-text-post={*}]}}
        \toprule
        & & \multicolumn{3}{c}{$\Delta$ ECE} & \multicolumn{3}{c}{$\Delta$ Brier Score} & \multicolumn{3}{c}{$\Delta$ Confidence} \\
        \cmidrule(lr){3-5} \cmidrule(lr){6-8} \cmidrule(lr){9-11}
        & Model & {P(True)} & {Verb. (\%)} & {Ling.} & {P(True)} & {Verb. (\%)} & {Ling.} & {P(True)} & {Verb. (\%)} & {Ling.} \\
        \midrule

        \multirow{7}{*}{\rotatebox[origin=c]{90}{\textbf{GSM8K}}} 
        & Llama 3.1 (8B) & 0.150* & 0.241* & 0.287* & 0.127* & 0.267* & 0.348* & 0.170* & 0.302* & 0.289* \\
        & Llama 3.1 (70B) & 0.248* & 0.123* & 0.091* & 0.187* & 0.124* & 0.103* & 0.366* & 0.170* & 0.038* \\
        & Qwen3 (4B) & 0.294* & 0.093* & 0.221* & 0.282* & 0.094* & 0.233* & 0.363* & 0.100* & 0.239* \\
        & Qwen3 (30B) & 0.011 & 0.003 & 0.009* & 0.014 & 0.003* & 0.007 & 0.085* & 0.007* & 0.011 \\
        & Gemma 3 (4B) & 0.143* & 0.012* & 0.204* & 0.155* & 0.014* & 0.228* & 0.162* & 0.012* & 0.204* \\
        & Gemma 3 (27B) & 0.004 & 0.009 & 0.020* & 0.005 & 0.006* & 0.007 & 0.022 & 0.016* & 0.031* \\
        \cmidrule(lr){2-11}
        & Average & 0.142 & 0.080 & 0.139 & 0.128 & 0.085 & 0.154 & 0.195 & 0.101 & 0.135 \\
        
        \midrule 

        \multirow{7}{*}{\rotatebox[origin=c]{90}{\textbf{TruthfulQA}}} 
        & Llama 3.1 (8B) & 0.053* & 0.038* & 0.078* & 0.048* & 0.035* & 0.035 & 0.109* & 0.193* & 0.283* \\
        & Llama 3.1 (70B) & 0.046* & 0.074* & 0.051* & 0.048* & 0.086* & 0.051* & 0.068* & 0.012 & 0.008* \\
        & Qwen3 (4B) & 0.087* & 0.029* & 0.164* & 0.089* & 0.024* & 0.070* & 0.079* & 0.036* & 0.138* \\
        & Qwen3 (30B) & 0.003 & 0.097* & 0.044* & 0.001 & 0.101* & 0.054* & 0.093* & 0.108* & 0.137* \\
        & Gemma 3 (4B) & 0.092* & 0.115* & 0.090* & 0.092* & 0.117* & 0.067* & 0.018* & 0.098* & 0.051* \\
        & Gemma 3 (27B) & 0.146* & 0.027* & 0.055* & 0.145* & 0.045* & 0.052* & 0.199* & 0.034* & 0.034* \\
        \cmidrule(lr){2-11}
        & Average & 0.071 & 0.063 & 0.080 & 0.070 & 0.068 & 0.055 & 0.094 & 0.080 & 0.109 \\

        \midrule 

        \multirow{7}{*}{\rotatebox[origin=c]{90}{\begin{tabular}{@{}c@{}}\textbf{MMLU} \\ \textbf{(Open-ended)}\end{tabular}}}
        & Llama 3.1 (8B) & 0.031* & 0.254* & 0.330* & 0.044* & 0.245* & 0.388* & 0.061* & 0.296* & 0.342* \\
        & Llama 3.1 (70B) & 0.046* & 0.013 & 0.013 & 0.046* & 0.002 & 0.020 & 0.042* & 0.029 & 0.028* \\
        & Qwen3 (4B) & 0.064* & 0.100* & 0.179* & 0.060* & 0.136* & 0.224* & 0.077* & 0.105* & 0.163* \\
        & Qwen3 (30B) & 0.002 & 0.232* & 0.242* & 0.006 & 0.241* & 0.225* & 0.006 & 0.256* & 0.283* \\
        & Gemma 3 (4B) & 0.176* & 0.138* & 0.232* & 0.180* & 0.149* & 0.225* & 0.203* & 0.142* & 0.246* \\
        & Gemma 3 (27B) & 0.056* & 0.152* & 0.092* & 0.055* & 0.150* & 0.089* & 0.063* & 0.161* & 0.113* \\
        \cmidrule(lr){2-11}
        & Average & 0.063 & 0.148 & 0.181 & 0.065 & 0.154 & 0.195 & 0.075 & 0.165 & 0.196 \\

        \bottomrule
    \end{tabular}
    \caption{Difference in ECE, Brier score, and raw confidence between answers provided by the assistant (model itself) and the user on GSM8K, TruthfulQA, and open-ended MMLU. Values represent $\Delta = \text{Assistant} - \text{User}$, so positive values indicate higher metrics for the assistant. * indicates significant difference ($p < 0.01$) according to the Wilcoxon signed-rank test (for Brier and confidence) or paired bootstrap resampling test (for ECE).}
    \label{tab:other_tasks}
\end{table*}

\subsubsection{Total Confidence}

\begin{figure}[t]
    \centering
    \includegraphics[width=\linewidth]{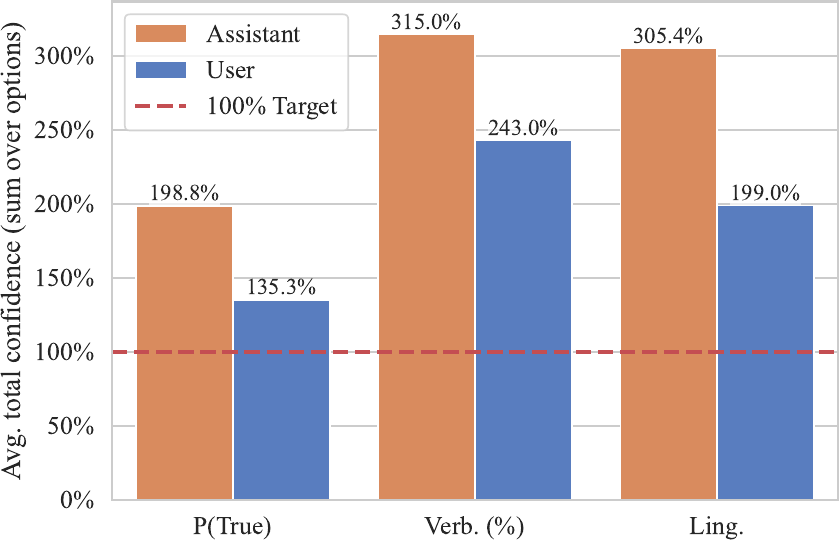}
    \caption{Average total confidence summed across all four options for each question, depending on whether the answer is provided by the assistant or the user. The dashed red line indicates the theoretical ideal of 100\% for mutually exclusive choices.}
    \label{fig:total_confidence}
\end{figure}

Since we experiment with a multiple-choice setting, the model's total confidence across the four options for a single question should ideally add up to 100\%.
Figure~\ref{fig:total_confidence} shows the total confidence (sum of confidences assigned to the four options) given by the model -- averaged across all models -- depending on whether the answer is provided by the assistant or the user.
We observe that, with all confidence estimation methods, the average total confidence is always higher than 100\%. This further indicates that LLMs are generally overconfident and are not consistent when assigning confidence to mutually exclusive options.
Additionally, the total confidence is much higher when the answer is provided by the assistant (ranging from roughly 198\% to 315\%) compared to the user (ranging from roughly 135\% to 243\%), reinforcing our hypothesis that LLMs are overconfident in their \emph{own} answers.

\subsubsection{Generalization to Other Tasks}
\label{sec:other_tasks}
So far, our analysis has focused on multiple-choice questions from the MMLU dataset, where the model is evaluated over a closed set of options. While this format enables precise calibration measurement, one could argue that explicitly listing candidate answers affects confidence behavior. To test whether our findings generalize beyond this setup, we extend the evaluation to GSM8K \citep{cobbe2021gsm8k}, TruthfulQA \citep{lin2022truthfulqa}, and an open-ended version of MMLU without predefined options.
These datasets cover diverse challenges: mathematical reasoning with open numeric answers (GSM8K), robustness to common misconceptions (TruthfulQA), and free-form factual answering (open-ended MMLU).
As shown in Table~\ref{tab:other_tasks}, the same pattern holds across all tasks: models are consistently better calibrated and less overconfident when judging answers framed as user input rather than as their own output.
On GSM8K, self-generated answers yield up to 19.5\% higher confidence and a 14.2\% increase in ECE.
On TruthfulQA, the confidence gap reaches up to 10.9\% with similarly higher calibration error.
Open-ended MMLU shows the same behavior, with up to 19.6\% higher confidence and 18.1\% higher ECE for assistant-framed answers.
Together, these results show that the ownership bias is not a multiple-choice artifact but a broader phenomenon across domains and answer formats.

\subsubsection{Generalization to Proprietary Models}
\label{sec:gpt5_results}
To test whether this behavior also appears in proprietary models, we further run an experiment with GPT-5.2 using MMLU.
Table~\ref{tab:gpt5_results} shows the same directional trend observed in open-weight models: assistant-framed answers produce higher ECE, Brier score, and raw confidence than user-framed answers across all three confidence elicitation methods.
ECE and confidence differences are statistically significant for all methods, while Brier score also degrades and reaches significance for the linguistic method.
These results indicate that ownership bias extends beyond open-weight LLMs and, thus, reinforce the generality of our findings as well as the practical relevance of our inference-time mitigation.

\begin{table}
    \small
    \centering
    \begin{tabular}{l *{3}{S[table-format=1.3, table-space-text-post={*}]}}
        \toprule
        Method & {$\Delta$ ECE} & {$\Delta$ Brier} & {$\Delta$ Confidence} \\
        \midrule
        P(True) & 0.077* & 0.042 & 0.076* \\
        Verb. (\%) & 0.087* & 0.047 & 0.112* \\
        Ling. & 0.113* & 0.095* & 0.222* \\
        \bottomrule
    \end{tabular}
    \caption{Difference in ECE, Brier score, and raw confidence between answers provided by the assistant (model itself) and the user for GPT-5.2 on MMLU. Values represent $\Delta = \text{Assistant} - \text{User}$, so positive values indicate higher metrics for the assistant. * indicates significant difference ($p < 0.01$) according to the Wilcoxon signed-rank test (for Brier and confidence) or paired bootstrap resampling test (for ECE).}
    \label{tab:gpt5_results}
\end{table}

\section{Conclusion}
In this work, we investigate the specific impact of the chat template on the miscalibration of instruction-tuned LLMs.
While we find that the chat format itself is not the main driver of miscalibration, it plays a critical role in how confidence is perceived: models exhibit an inherent ``ownership bias,'' showing significantly higher confidence in their own responses than in identical answers provided by a user, regardless of correctness.
Building on this insight, we propose a simple inference-time strategy that turns the chat format into an advantage: framing the model's answer as a user input during confidence elicitation. This method significantly reduces overconfidence and improves calibration across diverse benchmarks, narrowing the gap between base and instruction-tuned models. Our findings suggest that reliable confidence estimation requires decoupling the generation of an answer from its evaluation, forcing the model into a more objective ``observer'' role.

\section*{Limitations}
Our findings highlight the existence of overconfidence in LLMs' \emph{own} answers and offer a simple yet effective strategy to reduce it. However, there are several limitations to our study.

First, although we extend our analysis to GPT-5.2 (Section~\ref{sec:gpt5_results}), the majority of our experiments focus on open-weight LLMs due to the high cost of experiments and our commitment to open science and reproducibility. While our results cover a range of sizes, families, and include a proprietary model, we cannot fully guarantee that the same overconfidence behavior persists to the same degree in all proprietary, closed-source models where the post-training recipes (RLHF strategies and data mixtures) may differ.

Moreover, our proposed method -- framing model outputs as user inputs -- is an inference-time mitigation strategy. While it reduces miscalibration and narrows the gap between base and instruction-tuned models by leveraging the chat format, it does not alter the model weights or address the root causes of overconfidence introduced during the alignment process.

Finally, while we extended our analysis beyond multiple-choice questions to other tasks, our scope remains limited to objective question answering. It remains an open question whether similar overconfidence occurs in more subjective tasks (e.g., open-ended generation), where ``correctness'' is less clearly defined and confidence estimation is more ambiguous.

\section*{Acknowledgments}
We thank the anonymous reviewers for their helpful feedback.
This work was supported by the Carl Zeiss Foundation through the MAINCE project (grant number P2022-08-009).

\bibliography{refs}

\clearpage

\appendix

\section{Confidence Elicitation Prompts}
\label{app:prompts}

\begin{figure}[h]
    \centering
    \includegraphics[width=\linewidth]{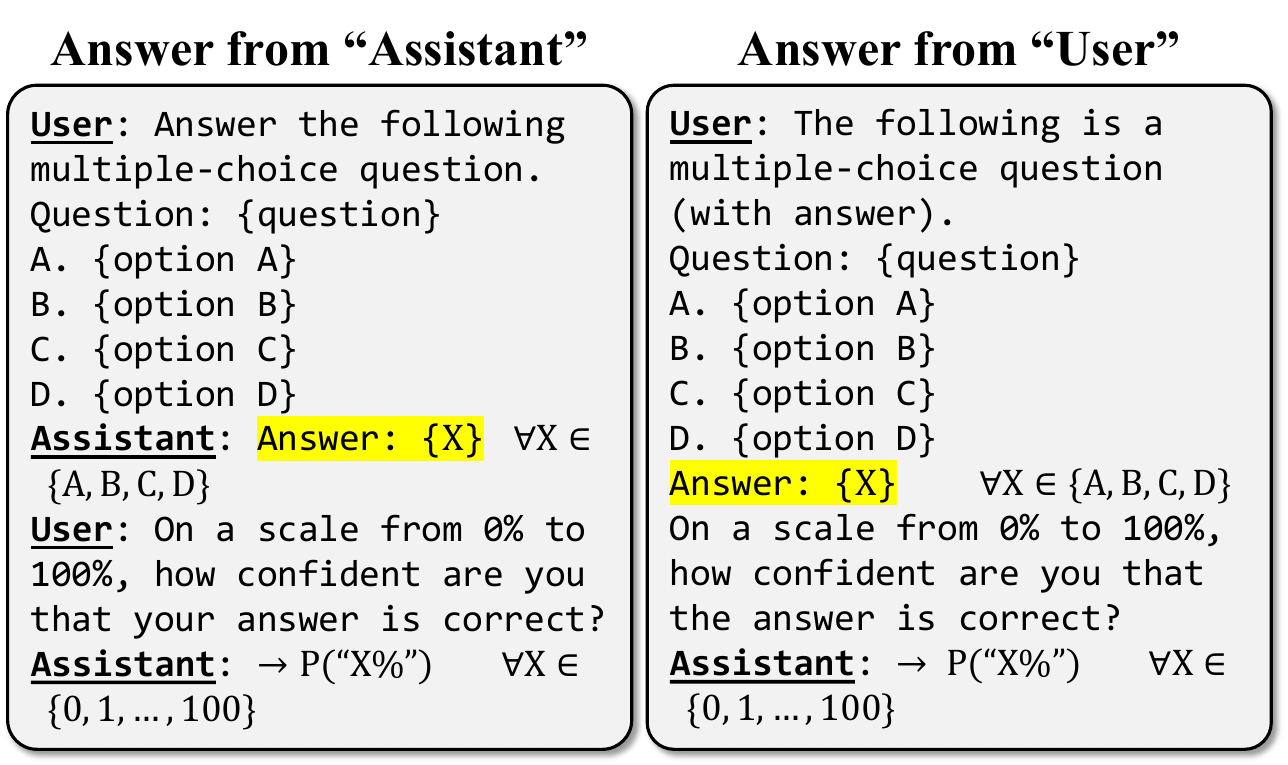}
    \caption{Prompts used to measure confidence in an answer provided by the model itself (left) and by the user (right) using the Verbalized Percentage method.}
    \label{fig:prompts_verbal_percentage}
\end{figure}

\begin{figure}[h]
    \centering
    \includegraphics[width=\linewidth]{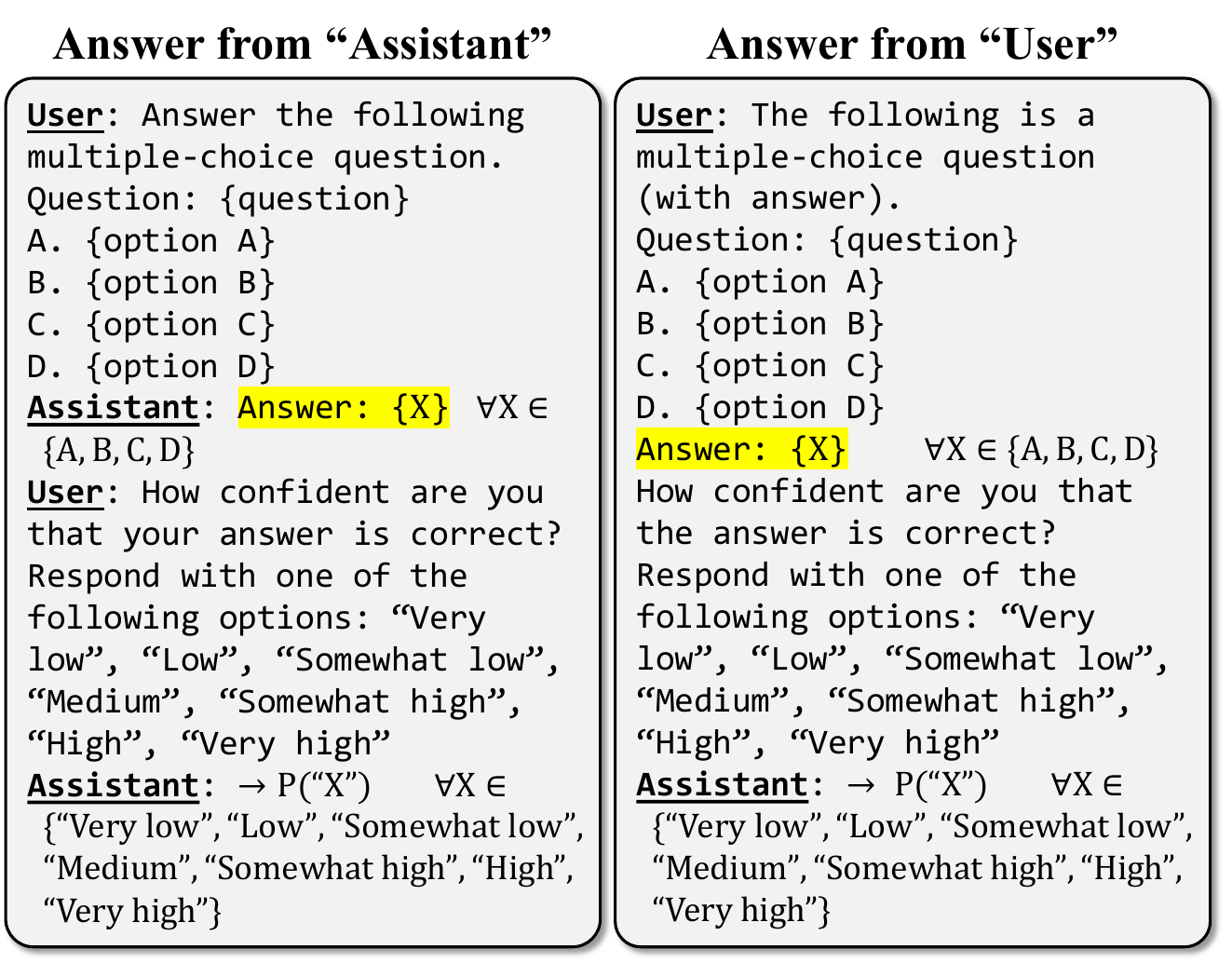}
    \caption{Prompts used to measure confidence in an answer provided by the model itself (left) and by the user (right) using the Verbalized Linguistic method.}
    \label{fig:prompts_verbal_linguistic}
\end{figure}

Figure~\ref{fig:prompts_p_true} in the main text shows the prompts used to measure confidence in an answer provided by the model itself (assistant) and by the user using the P(True) method.
Here, we provide the full prompts used for the other two confidence estimation methods: Verbalized Percentage (Figure~\ref{fig:prompts_verbal_percentage}) and Verbalized Linguistic (Figure~\ref{fig:prompts_verbal_linguistic}).

In all cases, the model's confidence is extracted from the completion with the highest log-probability among the options. To account for label length bias (where shorter labels are favored), we apply Normalized Contextual Calibration \citep{sanz-guerrero2025ncc} by subtracting the log-probability of each option in a neutral context (without the question) from the log-probability obtained in the original prompt.

Regarding the prompts used in open-ended tasks (Section~\ref{sec:other_tasks}), we use the same templates as in the multiple-choice setting, but we replace the list of options with a single answer (the one provided by the assistant or the user, depending on the condition) and ask the model to judge its confidence. This represents a standard closed-book, open-ended, zero-shot question answering scenario~\citep{shaier2025askingagain}.

\section{Full Results}
\label{app:full_results}

Table~\ref{tab:final_results} shows the full calibration results (ECE and Brier score) on MMLU when explicitly asking for confidence using the three confidence elicitation methods. For each model, we report results for the base version (no instruction-tuning), the instruction-tuned version without chat format, and the instruction-tuned version with chat format using both answer positions (assistant and user).
This Table provides a more detailed view of the gap bridged by our proposed method between base and instruction-tuned models.
In some cases, presenting the answer as part of the user message in the chat template not only recovers the calibration of the base model but even surpasses it.

\begin{table*}
    \centering
    \small
    \begin{tabular}{lccc *{6}{S[table-format=1.4]}}
        \toprule
         & & & & \multicolumn{3}{c}{ECE (↓)} & \multicolumn{3}{c}{Brier (↓)} \\
        \cmidrule(lr){5-7} \cmidrule(lr){8-10}
        Model & IT & Chat & Answer & {P(True)} & {Verb. (\%)} & {Ling.} & {P(True)} & {Verb. (\%)} & {Ling.} \\
        \midrule
        \multirow{4}{*}{Llama 3.1 (8B)} & \xmark & \xmark & N/A & 0.1910 & \bfseries 0.3483 & 0.3161 & 0.1930 & \bfseries 0.3174 & 0.2872 \\
         & \cmark & \xmark & N/A & 0.2796 & 0.6490 & 0.3767 & 0.2315 & 0.5992 & 0.3252 \\
         & \cmark & \cmark & Assistant & 0.2329 & 0.5133 & 0.4414 & 0.2236 & 0.4500 & 0.3707 \\
         & \cmark & \cmark & User & \bfseries 0.1609 & 0.4373 & \bfseries 0.2064 & \bfseries 0.1916 & 0.3790 & \bfseries 0.1547 \\
        \midrule
        \multirow{4}{*}{Llama 3.1 (70B)} & \xmark & \xmark & N/A & 0.0267 & \bfseries 0.1733 & \bfseries 0.2745 & \bfseries 0.0992 & \bfseries 0.1405 & \bfseries 0.1983 \\
         & \cmark & \xmark & N/A & 0.0744 & 0.2135 & 0.3151 & 0.1030 & 0.1739 & 0.2288 \\
         & \cmark & \cmark & Assistant & 0.1130 & 0.4137 & 0.4511 & 0.1825 & 0.2500 & 0.3818 \\
         & \cmark & \cmark & User & \bfseries 0.0220 & 0.3937 & 0.4261 & 0.1155 & 0.2280 & 0.3528 \\
        \midrule
        \multirow{4}{*}{Qwen3 (4B)} & \xmark & \xmark & N/A & \bfseries 0.1635 & \bfseries 0.4539 & 0.4123 & \bfseries 0.1732 & 0.4182 & 0.3598 \\
         & \cmark & \xmark & N/A & 0.2268 & 0.5850 & 0.5021 & 0.2186 & 0.5544 & 0.4241 \\
         & \cmark & \cmark & Assistant & 0.2140 & 0.6535 & 0.5445 & 0.2145 & 0.6026 & 0.4762 \\
         & \cmark & \cmark & User & 0.2120 & 0.4605 & \bfseries 0.1335 & 0.2035 & \bfseries 0.3756 & \bfseries 0.0602 \\
        \midrule
        \multirow{4}{*}{Qwen3 (30B)} & \xmark & \xmark & N/A & \bfseries 0.1326 & 0.4037 & 0.3589 & \bfseries 0.1415 & 0.3377 & 0.2806 \\
         & \cmark & \xmark & N/A & 0.2340 & 0.5835 & 0.4565 & 0.1829 & 0.5001 & 0.3650 \\
         & \cmark & \cmark & Assistant & 0.1935 & 0.6759 & 0.5674 & 0.1902 & 0.6368 & 0.4916 \\
         & \cmark & \cmark & User & 0.1485 & \bfseries 0.3289 & \bfseries 0.1314 & 0.1502 & \bfseries 0.2828 & \bfseries 0.0636 \\
        \midrule
        \multirow{4}{*}{Gemma 3 (4B)} & \xmark & \xmark & N/A & \bfseries 0.0209 & \bfseries 0.4558 & 0.3452 & \bfseries 0.1882 & \bfseries 0.3930 & 0.3062 \\
         & \cmark & \xmark & N/A & 0.5779 & 0.7132 & 0.5418 & 0.5645 & 0.6929 & 0.4731 \\
         & \cmark & \cmark & Assistant & 0.6546 & 0.7130 & 0.5477 & 0.6517 & 0.6953 & 0.4959 \\
         & \cmark & \cmark & User & 0.4026 & 0.5920 & \bfseries 0.2797 & 0.3987 & 0.5463 & \bfseries 0.2499 \\
        \midrule
        \multirow{4}{*}{Gemma 3 (27B)} & \xmark & \xmark & N/A & \bfseries 0.0428 & 0.5213 & 0.3356 & \bfseries 0.1798 & 0.4465 & 0.2942 \\
         & \cmark & \xmark & N/A & 0.2395 & 0.5465 & 0.5256 & 0.2366 & 0.4785 & 0.4520 \\
         & \cmark & \cmark & Assistant & 0.3179 & 0.6771 & 0.4560 & 0.3171 & 0.6461 & 0.3813 \\
         & \cmark & \cmark & User & 0.1919 & \bfseries 0.3621 & \bfseries 0.2650 & 0.1911 & \bfseries 0.3011 & \bfseries 0.2063 \\
        \midrule
        \midrule
        \multirow{4}{*}{\textit{Average}} & \xmark & \xmark & N/A & \bfseries 0.0963 & \bfseries 0.3927 & 0.3404 & \bfseries 0.1625 & \bfseries 0.3422 & 0.2877 \\
         & \cmark & \xmark & N/A & 0.2720 & 0.5485 & 0.4530 & 0.2562 & 0.4998 & 0.3780 \\
         & \cmark & \cmark & Assistant & 0.2877 & 0.6078 & 0.5014 & 0.2966 & 0.5468 & 0.4329 \\
         & \cmark & \cmark & User & 0.1897 & 0.4291 & \bfseries 0.2404 & 0.2084 & 0.3521 & \bfseries 0.1813 \\
        \bottomrule
    \end{tabular}
    \caption{\textbf{Full calibration results} (ECE and Brier score) for all models in different settings: (1) base model without instruction tuning (``IT'') and without chat format (``Chat''); (2) instruction-tuned model without chat format; (3) instruction-tuned model with chat format answering as assistant; and (4) instruction-tuned model with chat format answering as user. Best results for each model and metric are highlighted in bold.}
    \label{tab:final_results}
\end{table*}

\end{document}